# A Unified Training Process for Fake News Detection based on Fine-Tuned BERT Model

Vijay Srinivas Tida, Sonya Hsu, Xiali Hei

*Abstract*—An efficient fake news detector becomes essential as the accessibility of social media platforms increases rapidly. Previous works mainly focused on designing the models solely based on individual datasets and might suffer from degradable performance. Therefore, developing a robust model for a combined dataset with diverse knowledge becomes crucial. However, designing the model with a combined dataset requires extensive training time and sequential workload to obtain optimal performance without having some prior knowledge about the model's parameters. The presented work here will help solve these issues by introducing the unified training strategy to have a base structure for the classifier and all hyperparameters from individual models using a pre-trained transformer model. The performance of the proposed model is noted using three publicly available datasets, namely ISOT and others from the Kaggle website. The results indicate that the proposed unified training strategy surpassed the existing models like Random Forests, Convolutional Neural Networks (CNN), Long Short Term Memory (LSTM), etc., with 97% accuracy and achieved the F1 score of 0.97. Furthermore, there was a significant reduction in training time by almost 1.5 to 1.8 $\times$ by removing words lower than three letters from the input samples. We also did extensive performance analysis by varying the number of encoder blocks to build compact models and trained on the combined dataset. We justify that reducing encoder blocks resulted in lower performance from the obtained results.

## I. INTRODUCTION

Fake news can spread wrong information to the public, which causes biased opinions, changes real-world perspectives, and provokes violence. 62% of US individuals have access to social media platforms, so it requires much attention to avoid unforeseen circumstances resulting from spreading fake information [1, 2]. For example, during the 2016 US elections [3] and the COVID pandemic [4], much misinformation spread across the Internet and made some individuals trust it. This belief resulted in casting their votes for the wrong person [5] and facing severe health problems through home treatment [6]. Reputable media delivers news to the public that can be most trustworthy. But mainly, minor news media provides fake news to become popular instead of giving the facts to the viewers [7]. The spreading of untrusted information can cause severe problems like buying unworthy products by consumers, fluctuations in the stock market, the mental health of individuals, etc [8, 9, 10]. Hence, there is a need for the public to have an efficient fake news detector to solve the problems mentioned above to some extent.

Vijay Srinivas Tida, Sonya Hsu, Xiali Hei are with the School of Computing and Informatics, University of Louisiana at Lafayette, Lafayette, LA 70503, USA (e-mail: vijaysrinivas.tida1@louisiana.edu, hsiuyueh.hsu@louisiana.edu, xiali.hei.@louisiana.edu).

Fake news detection models can be classified into content-based and context-based learning methods. For content-based learning, the approaches used in [11, 12, 13] deal with extracting the features related to writing style and the information from fake news articles. The other methods in [14, 15] use only linguistic features to capture the writing style to identify the spreading of fake news that happens similarly to actual stories. However, for efficient fake news detection using content-based features might not be sufficient alone. Thus, we also need context-based approaches used in [12, 16, 17] that have the relationship between the news article and the user. Most of the above methods use word embedding models like word2Vec, GloVe, Term Frequency-Inverted Document Frequency(TF-IDF), etc.

The popular existing fake news detection models usually will be trained on a single dataset, and they might not perform well in actual world conditions due to the limited information available from a specific domain. Moreover, the datasets are required to combine for the given task from various fields to have rich knowledge [18]. This helps the model perform well when compared to the prior case. Consequently, there is a dire need to develop an efficient unified fake news detection model when different datasets are combined. Furthermore, the combined dataset has varied data distribution and extensive feature selection, which covers more domains [19]. Therefore, training a model with the combined dataset having these characteristics is a hectic task. Hence, an efficient training strategy is needed for the designed model such that it does not have low performance and can extract rich information from the various input samples [20].

Previous work proposed by Ahmad *et al.* [21] showed that ensemble machine learning models trained on a combined dataset perform better than deep learning models. The performance of machine learning and deep learning models for various applications like computer vision, cybersecurity, natural language processing, etc., are illustrated in [22, 23, 24, 25]. Usually, the machine learning models tend to saturate after a particular number of data samples [26], but in [21], they performed well for fake news classification with a combined dataset. On the other hand, deep learning models using the combined dataset showed lower performance among the proposed models in [21]. The main reason for performance issues is that deep models do not have prior knowledge about the assigned task as they are trained from scratch.

Generally, deep learning models perform better than machine learning models due to the increasing data samples [26]. This paper tries to address these issues with the help of a unified training process using a transfer learning approach to obtain a finetuned transformer model with better performance than machine learning approaches. The proposed method allows the performance loss within acceptable limits compared to the previous works on trained individual models. This relaxation helps to get the base classifier structure for the unified model with similar hyperparameters by having three same models. Thus, the unified model can be trained on the combined dataset with the obtained base structure and hyperparameters. We are the first to propose the unified training process by intentionally allowing some performance loss for individual models to get an optimal final model. Even though the original authors for the selected pretrained model [27] didn't recommend using input preprocessing, we purposefully did it to reduce training time by removing words of length less than three letters.

The significant contributions of this paper are as follows:

a. We propose a unified training process to achieve a model for efficient fake news detection that can train on a combined dataset. The unified training process involves obtaining the base classifier structure and the hyperparameters from the three individual models trained on three datasets separately.

b. We analyze the performance and the training time required when the input samples were fed into the unified model with and without preprocessing. Our experimental results indicate that models trained with input pre-preprocessing showed better performance and recorded a lower training time when compared to the model without input preprocessing.

c. We investigate the performance of the unified model by varying the number of encoder blocks. Our experimental results showed that reducing the number of encoder blocks decreases the model's performance.

The rest of the paper is organized as follows: Section II overviews previous research works on fake news detection applications. Then, Section III explains the proposed unified training process to obtain a model that can be trained on a combined dataset. After that, Section IV presents the results using the proposed approach, and Section V discusses the possible future research directions. Finally, Section VI concludes the proposed research work.

## II. LITERATURE REVIEW

Previous works can be grouped into machine learning, deep learning, and transfer-learning categories based on available approaches for fake news detection.

*1) Machine learning based approaches:* Ahmed *et al.* [28] used multiple machine learning models and obtained a higher accuracy of 92%. The authors combined TF-IDF as the feature extractor and Linear Support Vector Machine as the classifier. Shu *et al.* [29] extensively surveyed fake news detection models using textual and visual feature combinations to improve accuracy. The authors showed model-oriented approach performs well for counterfeiting news detection. Gilda *et al.* [30] used a dataset from Signal Media for fake news detection. The results indicated that the Stochastic Gradient Descent model fed the input from TF-IDF of bi-grams showed 77.2% of accuracy. Kaliyar *et al.* [31] proposed ensemble-based learning using the Gradient Boosting algorithm for multiclass classification of fake news and achieved the accuracy of 86%. Finally, Singh *et al.* [32] fed Linguistic Analysis, and Word Count (LIWC) features for Support Vector Machines (SVMs) showed 87% of accuracy.

Hakak *et al.* [33] proposed an ensemble-based machine learning model comprised of Decision Tree and Random Forest and achieved training and test accuracy of 99.8% and 44.15% for fake news detection, respectively. Zhou *et al.* [34] proposed a theory-driven model by analyzing content-based, and propagation-based methods showed an average accuracy of 88% with a 2-8% improvement over previous approaches. Ahmad *et al.* [21] used ensemble models for fake news detection. They trained the model by combining three datasets and achieved higher accuracy of 91% using the Random Forest approach.

For designing the machine learning models, the user needs to manually extract the features to train the model for each dataset separately. The extraction of features specifically for the proposed model leads to a biased model in which usage of the same preprocessing steps might not perform well for other datasets. Most existing machine learning models showed the best performance specific to a single dataset. When multiple datasets are combined, machine learning models might not work well as expected due to saturation problems after achieving a certain level of accuracy. Furthermore, increasing training data samples might not help improve the performance of the machine learning models.

*2) Deep learning based approaches:* Wang *et al.* [35] proposed a Bidirectional LSTM model to predict the outcomes of the given samples. The authors focused only on the data from the political domain with enhanced input features and achieved a higher accuracy of 27.4%. Karimi *et al.* [36] proposed a neural network model that combines Convolutional Neural Network (CNN) and LSTM methods with the accuracy of 38.81%. The proposed method showed an improvement in accuracy compared to [35]. Ghanem *et al.* [37] proposed a new model with different word embeddings and n-gram features for stance detection in fake articles and achieved about 48.80% of accuracy. O'Brien et al. [38] proposed a fake news detector model with the help of a black-box based deep learning model using capturing the input words for classification and achieved the accuracy of 93.50%. Kaliyar *et al.* [39] proposed FNDNet that uses deep convolutional neural network for fake news detection. The model consists of multiple convolutional 1D layers, which can extract multiple features from the given input samples achieving the accuracy of 98.36% with a less false positive rate of 0.59.

Goldani *et al.* [40] proposed a capsule neural network with static and non-static embeddings for fake news detection. The proposed model achieved 99.8% accuracy for fake news classification using the ISOT dataset [41]. Ruchansky *et al.* [14] proposed a hybrid deep learning model that uses Capture, Score, and Integrate modules with the help of Recurrent Neural Networks for fake news classification and achieved the accuracy of about 89.20%. Nasir *et al.* [42] proposed a hybrid CNN-RNN-based model and achieved 99% of accuracy. Monti *et al.* [43] proposed a geometric deep learning model with two-dimensional features having two convolutional layers, showing 92.7% of accuracy.

In the case of deep learning models, there might be an improvement over the existing best-performing machine learning models when training data samples are increased. The reason is that the deep learning model doesn't require the user to preprocess the input data samples. Instead, the features will be extracted automatically from the input samples during the training process. However, the existing approaches might suffer from low performance in real-world cases in which the knowledge gained through the specific dataset is limited. To avoid these problems, pre-trained models can fill this gap using a knowledge transfer approach known as transfer learning.

*3) Transfer learning based approaches:* Jwa *et al.* [44] used a Bidirectional Encoder Representation from Transformers(BERT) model and showed an increase at 0.14 of the F1 score. Here the authors made minor modifications to the existing pre-trained BERT model by replacing the Cross-Entropy loss function with Weighted Cross Entropy. Kaliyar *et al.* [45] proposed the FakeBERT model using only one dataset but showed 98.9% of accuracy. In addition, the proposed FakeBERT achieved fewer False Positives and False Negatives.

Cruz *et al.* [46] suggested fake news detection using Multitask Transfer Learning approach for the Filipino language and showed an increase in accuracy of 4-6% using the transfer learning approach. To compare the performance of the proposed model, the authors used the Siamese Network as the baseline. Palani *et al.* [47] proposed a hybrid model using capsule neural network (CapsNet) and BERT for fake news detection and showed an average accuracy of 92%. The CapsNet is used for information extraction from the visual images, and BERT is used for extracting the context-based textual features from the hybrid model. Finally, Blackedge *et al.* [48] analyzed different pre-trained models' performance for fake news detection with baseline machine learning and deep learning algorithms. The results indicate that the deBERTa performed better than Logistic Regression, Naive Bayes, Random Forest, LSTM, DistilBERT, and BERT models.

The above-proposed models are only trained on a single dataset and might not be helpful for real-world situations. To solve this problem to some extent, we proposed a unified model that can be trained using multiple datasets with the help of a transfer learning approach using transformers. Tida *et al.* [49] proposed the universal model for spam detection in which four datasets were fed to the model for training and achieved 97% accuracy. The primary strategy here is that individual models are intentionally made to perform lower to obtain the robust unified model. We follow a similar training process to get a powerful detection model for fake news classification.

## III. METHODOLOGY

### A. Model selection and its overview

*1) Model selection:* Deep learning models showed significant improvement for NLP applications over the past decade [50]. Among them, Long Short Term Memory(LSTM) showed better performance by taking the time instant into consideration [51]. However, the vanishing gradient problem in LSTM models arises when more extensive data sequences exist in the given input sample. Therefore, the maximum sequence length for the given input sample should be 200 to avoid this problem. Furthermore, computation time will also be longer due to sequential dependency during the training phase of the model [52]. Later, transformer models solved these problems by removing sequential dependency with position encoding and allowing the higher sequence length limits up to 512 [27, 53]. Finally, we selected Google's BERT model for the fake news detection task, which uses the transfer learning approach.

*2) Overview of selected pre-trained model design:* Google released various BERT transformer versions popularly used for applications like sequence-to-sequence modeling and classification tasks. Based on the application and scope of the problem, users can select the appropriate one from the available transformers. As the fake news detection task is considered a classification task and the scope of this task needs a compact structure that is case insensitive, we selected a base model with an uncased version. For classifying the given input samples, the output vector corresponding to the classifier part can be used and designed accordingly by keeping the pre-trained weights fixed.

Transformers generally contain both encoder and decoder units, whereas BERT consists of only the encoder part by discarding the decoder part. The presence of only the encoder part helps to reduce the complexity of the original structure, as the standard transformer feeds the hidden state information from the encoder to the decoder. Bidirectional means that the processing of the input sequence takes from both backward and forward directions. This process helps the model learn from both directions to predict the word in the given context more efficiently. The selected base BERT model contains 12 encoders with 110 million pre-trained weights. Since the selected finetuned model includes fewer trainable parameters, it can help reduce the training time. Using a semisupervised approach, the BERT transformer model was pretrained on two large databases: Wikipedia (2.5 million words) and book corpus (800 million words).

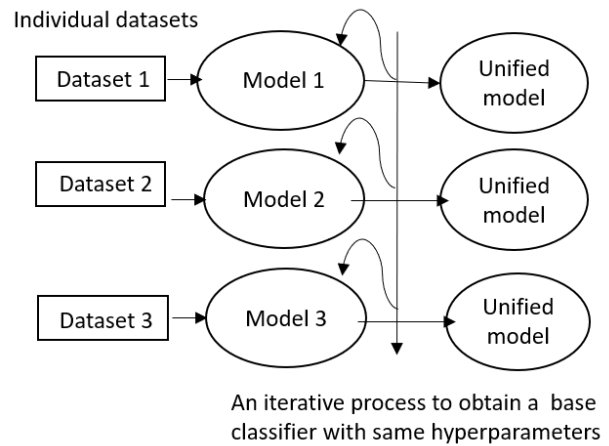

**Fig. 1: A unified training process for fake news detection trained on three individual datasets**

*B. A unified training process for the fake news detection*

The unified training process involves two phases, as seen in Fig. 1 and 3. In the first phase, we obtain a unified model structure by allowing the design constraints by training models separately on three datasets. The form of the base classifier obtained can be seen in Fig. 2. We will finetune the accepted unified model in the second phase by combining all the datasets. The detailed explanation of the unified training process is as follows:

*1) Constraints needed for the unified training process:* Hyperparameter tuning is made for all three individual models simultaneously to have a base classifier structure with acceptable performance. The acceptable performance means that the obtained models trained on individual datasets do not need to be the best-performing models. The difference between the accepted and best-performing previous models should not be significant. For the proposed fake news detection, we initially allowed a threshold of 1 to 10% difference in accuracy from the best-performing model during the training process. Later, hyperparameter tuning is made on individual models, helping to improve the performance. The final goal is to have an efficient base structure with common hyperparameters for the unified model to train on the combined dataset.

*2) Procedure for obtaining the unified model:* The first phase of the training process involves obtaining the unified model, as seen in Fig. 1. In this part, initially, we design the three models to have the best performance on the individual datasets. The designing process of individual models includes input preprocessing by removing words less than three let- ters before feeding into the pre-training model. Although the authors in [27] didn't recommend input preprocessing but doing this help reduce the computation load and the training time. Also, removing these words will not affect the performance since the model is designed for the classification task. However, this technique will not work for sequence-to-sequence applications as it requires a whole input data sample. Then after obtaining better performing individual models, we applied constraints and iteratively modified the model to have an efficient base classifier for all three models. Thus, the performance of the individual models falls about 10% on average from the best performing models. After hyperparameter tuning, the first model showed only a 0.04% decrease in accuracy, and the other two models performed well compared to the previous works. The final goal of the first phase is achieved by obtaining a unified model having the same structure with hyperparameters.

*3) Base classifier design:* Fig. 2 depicts the finalized model architecture for the classifier obtained after having common hyperparameters. First, the output vector of 768 from the BERT pre-trained model was used as input to the linear layer with 200 neurons. Next, the batch normalization layer with Rectified Linear Unit (ReLU) and the dropout layer with a dropout rate of 0.1 is added over the linear output. A similar structure is repeated once but changed to 150 neurons in the linear layer. At the final stage of the finetuned classifier, a linear layer is added, containing two neurons, and the activation function log softmax is applied to the output. The log softmax activation output will help classify whether the input sample is fake or real. After allowing the constraints, the final classifier was used for joint training after the finetuning process for individual models on the corresponding datasets.

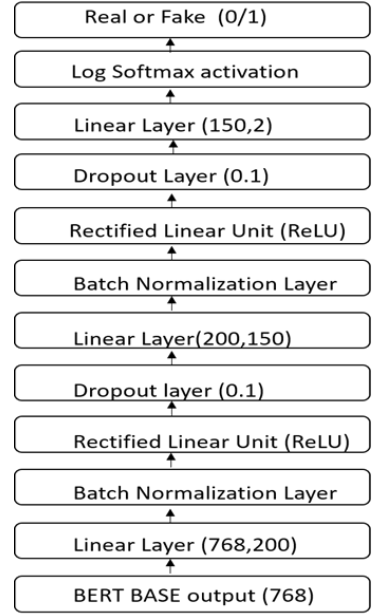

**Fig. 2: The base classifier design**

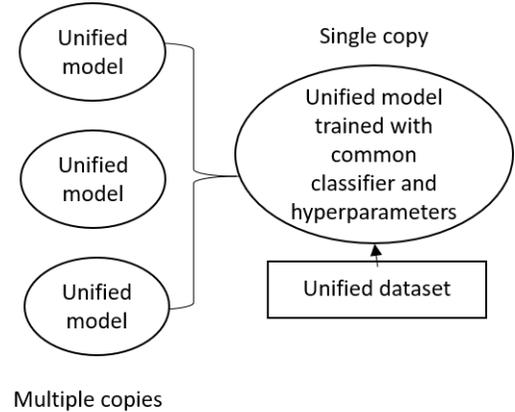

**Fig. 3: A joint training process for fake news detection trained on the combined dataset**

*4) The joint training process and the formation of the compact model:* The joint training process involves training a unified model obtained from phase 1, as seen in Fig. 3. The accepted base structure with hyperparameters will help reduce the time needed for the efficient model search for the combined dataset. However, training the model on a large dataset is tedious and might require a lot of time to obtain the best-performing model. Thus, the accepted structure with the corresponding hyperparameters is used as the initial step for training the model. Therefore, the joint training process involves training the unified model on the combined dataset to get the best-performing model.

The preprocessed input and the strategies applied to obtain common hyperparameters for the individual models reduced the unified model's sequence length from 300 to 175. Thus, decreasing the sequence length value helps the model to reduce the computation load, which shortens the training time. Therefore, removing the words with less than three letters will not affect the performance of the task and will help the model converge faster. Here the initial weights of the classifier design are chosen randomly without transferring the weight values from the unified models. The training process involves modifying hyperparameters with slight modifications to the classifier design. However, the model structure obtained from phase 1 showed better results for the fake news classification task without changing the base structure and hyperparameters. We only varied the batch size to analyze the performance and trained the model accordingly with the same classifier and hyperparameters. To make the model further compact, we analyzed the unified model's performance by changing the number of encoder blocks from the obtained unified model.

### C. The workflow

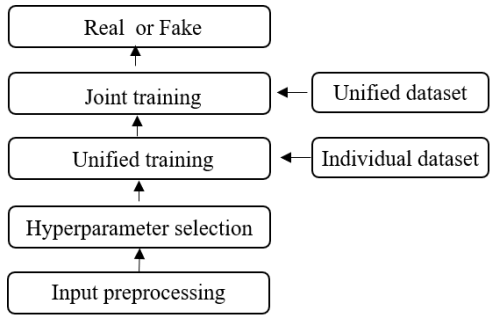

**Fig. 4: The workflow for the unified training process**

The workflow for the unified training process is shown in Fig. 4 is explained below:

1. The first step of the proposed unified approach for fake news detection involves removing the unnecessary words from the input data samples from the three individual datasets. The input preprocessing includes removing words whose length is less than three letters. These words don't have important information and are useless for classification tasks. Thus, removing these words helps the model train faster than the model with input preprocessing.

2. The preprocessed input samples are fed into the corresponding selected pre-trained models in the second step. Then, hyperparameters are initialized accordingly to start the finetuning process of the individual models. Then, the hyperparameters are chosen based on constraints to obtain the unified model structure. The details about constraints are explained clearly in the next step.

3. In the third step, the unified model is obtained by iteratively finetuning the three individual models using three datasets to have the same classifier structure with hyperparameters. The final goal of this step is to get the three similar classifier designs with hyperparameters having acceptable performance. The process of getting a unified model from this step can be referred to as the unified training process.

4. In the fourth step, the obtained unified model from the previous step is finetuned using a unified dataset with input preprocessing. Here the finetuning process starts with the hyperparameters obtained from the last step. After finetuning process, the unified model parameters are varied accordingly to get the best performing model. This process is called the joint training process.

Thus, in our case, the initial unified model performed better on the combined dataset, so there is no need to change the model parameters. Finally, with the help of a unified training approach, we achieved a robust fake news detection model. Therefore, any classification task can use the abovementioned process in NLP without restrictions.

## IV. RESULTS

### A. Dataset description

We used three publicly available datasets to evaluate the proposed model. The three datasets used in the proposed work are ISOT dataset [41] and the other two from the Kaggle website [54, 55]. The datasets contain both real and fake news articles from various domains. The actual samples are obtained from trusted sources that contain truthful information. On the other hand, the rumors and the media that do not have trusted information are being spread through multiple media platforms represented as fake samples. These samples are generally labeled manually with fact-checking websites like politifact.com, snopes.com, etc. The detailed description of the datasets can be clearly explained below:

The first dataset used for our analysis is the "ISOT Dataset," containing real and fake samples extracted from online sources. For evaluation purposes, this dataset can be termed dataset1. The real samples were gathered from the website named reuters.com, which is one of the popular and trusted news websites, while fake samples were collected from different websites flagged by politifact.com. The dataset consists of 21,417 real articles and 23,481 fake articles. In total, 44,898 articles are present in the entire dataset. The data samples from this dataset are mainly related to political news.

The second dataset used for the evaluation is from the Kaggle website. For evaluation, this dataset can be termed dataset2. This dataset contains 20,386 samples used for training purposes, and 5,126 samples are used for testing purposes. The samples include information from various domains on the internet. The articles cover a wide range of topics like entertainment and sports, not only related to politics [54]. Thus both real and fake samples contain broad information from different domains, which have a rich amount of knowledge compared to the dataset1.

The third dataset is also accessed from the Kaggle website [55]. In this paper, we refer to this dataset as dataset3. It contains a total of 3,352 samples. Again, various authenticated online sources like CNN, the New York Times, Reuters, etc., are used to label the real samples. In this dataset, the samples will cover a wide range of topics, including politics, entertainment, and sports. Similarly, untrusted websites help to mark the fake samples in the dataset.

The samples from all three datasets are combined to get the larger dataset and cover termed unified/combined dataset in our proposed work. Since the articles vary in real-world scenarios, this combined dataset will help to evaluate the performance in which the samples cover a rich range of domains in a single dataset. The designed model is a supervised algorithm in which we labeled '0' for real news samples and '1' for fake news samples: the information obtained from the samples and the encoded output labels from the samples were used to train and evaluate the proposed model's performance.

*B. Improvement techniques*

Model performance is improved with the help of Xavier initialization [56] and an Adam optimizer with a learning rate of 0.003 [57]. The addition of these two methods for training helps the model converge faster. The main goal of adding Xavier's initialization is to initialize the weights such that the variance of the activations will be the same across each layer. Adam optimizer is considered the better optimization algorithm with faster computation time, and tuning it requires fewer parameters. Finally, to avoid an exploding gradient problem, we use the gradient clipping technique with a clip value of 1 [58]. The model weights with the best validation accuracy are used for subsequent epochs during the training phase.

*C. The training process and performance analysis of the proposed model for three datasets*

We made the train-test distribution 80:20 so that the designed model can be compared easily with the previous works. The final goal of this process is to develop a final unified model to obtain higher accuracy without showing bias towards the samples from any dataset. The performance analysis for the individual models using the three datasets is made by varying minibatch size with the help of a base classifier and hyperparameters obtained by applying the constraints. The obtained results for the unified model using three datasets separately by training with 50 epochs can be seen in Table I. The results show that dataset1, dataset2, and dataset3 showed higher accuracy and F1-score with the minibatch sizes of 32, 64, and 128, respectively. Therefore, the trained individual models using a unified strategy showed acceptable performance without being significantly different from the best-performing model.

*D. Performance analysis of the proposed model for three datasets with previous works*

The comparison of obtained results with previous works for dataset1 can be seen in Table II. Our finalized model showed 99.4% of accuracy and suffered a performance loss of 0.6% from the best performing model proposed by Hakkak *et al.* [33]. The obtained accuracy indicated that there is not much performance loss compared to the best-performing model. This sacrifice of individual accuracy can be considered a trade-off to get a better-unified model. From Table III, the obtained model trained on dataset2 performed well with 97% of accuracy. However, machine learning models like Random Forest and Decision Trees showed the accuracy of 85% and 92%, respectively, and deep learning models like LSTM and Bi-LSTM models showed worse performance. Similarly, the same model for dataset3 showed 99% of accuracy, which can be seen in Table IV. The unified model performed well on these two datasets and showed higher accuracy than previous works.

*E. Performance analysis of the unified model using a combined dataset with and without input preprocessing*

In this part, we will analyze the importance of input preprocessing for the unified model regarding the training time and the performance metrics. Table V indicates the unified model's performance without preprocessing input data samples trained from the model structure obtained from individual models. For this model, the hyperparameter named sequence length is set to 200 according to the distribution of samples. Results indicate the highest performance of 95% accuracy with an F1-score of 0.9471 is achieved for a minibatch size of 32. The average training time for the model without preprocessed input data was about 22 hours using RTX 8000 GPU due to the more sequence length value. Table VI indicates the unified model's performance by giving preprocessed input data samples as input for the unified model. The preprocessed input data and the constraints helped reduce the sequence length to 120 from 200. The training time for the model was significantly reduced to 12 hours from 22 hours by a simple modification to the input samples and achieved the highest performance. Results showed the highest performance for minibatch size of 64 with 97% of accuracy and a corresponding F1-score of 0.97. The results also indicate that input preprocessing helps increase performance with reduced training time.

*F. Comparison of results obtained from the unified model for four datasets with previous works*

Table VII shows the accuracy, F1-score, precision, and recall values for the existing models obtained from [21] compared with the proposed approach. The authors used various machine learning and deep learning algorithms for fake news classification and showed Random Forest model performed well in most cases. For dataset1, the Random Forest algorithm and the proposed model showed higher performance in terms of accuracy, F1-score, and precision metrics but obtained less recall value. However, the proposed model showed similar results to the LSVM algorithm [64] when dataset1 is considered. Deep learning models like CNN [35] and Bi-LSTM [35] showed less performance than all other approaches.

Similarly, for dataset2, our proposed model outperformed all other models in all four metrics with 97% of accuracy for a batch size of 64. The second best performing model was Boosting classifiers with AdaBoost and XGBoost showed 94% of accuracy. However, for dataset2, the Random Forest algorithm doesn't perform well as like dataset1. But deep learning models like CNN [35] and Bi-LSTM [35] also showed limited performance on dataset2. Also, for dataset3, our proposed model performed well compared with the highest accuracy of 99% with 128 as batch size. LSVM [64] model showed better performance next to our proposed model, with 96% of accuracy. However, other deep learning models also showed poor performance on dataset3. The main reasons for the limited performance of deep learning models might need more training samples, and there is no prior knowledge to handle the classification task efficiently.

| Minibatch size | Dataset1 | | | | Dataset2 | | | | Dataset3 | | | |
|---|---|---|---|---|---|---|---|---|---|---|---|---|
| | Accuracy | Precision | Recall | F1-score | Accuracy | Precision | Recall | F1-score | Accuracy | Precision | Recall | F1-socre |
| 16 | 0.98 | 1 | 0.96 | 0.9793 | 0.97 | 0.99 | 0.95 | 0.9712 | 0.98 | 0.99 | 0.97 | 0.9810 |
| 32 | 0.99 | 0.99 | 0.99 | 0.9883 | 0.97 | 0.97 | 0.97 | 0.9721 | 0.99 | 0.99 | 0.98 | 0.9838 |
| 64 | 0.98 | 0.96 | 0.99 | 0.9748 | 0.97 | 0.96 | 0.99 | 0.9743 | 0.99 | 0.99 | 0.97 | 0.9837 |
| 128 | 0.98 | 0.99 | 0.98 | 0.9824 | 0.95 | 0.91 | 0.99 | 0.9511 | 0.99 | 0.99 | 0.98 | 0.9893 |
| 256 | 0.98 | 0.99 | 0.96 | 0.9764 | 0.96 | 0.99 | 0.94 | 0.9634 | 0.99 | 0.99 | 0.98 | 0.9787 |
| 512 | 0.98 | 0.99 | 0.98 | 0.9824 | 0.96 | 0.99 | 0.93 | 0.9614 | 0.98 | 0.97 | 0.99 | 0.9814 |
| 1024 | 0.98 | 0.96 | 0.99 | 0.9758 | 0.97 | 0.95 | 0.98 | 0.9665 | 0.99 | 0.99 | 0.98 | 0.9865 |

**TABLE I:** Accuracy, Precision, Recall, and F1-score for the dataset1, dataset2, and dataset3 by varying the batch size

| Authors | Accuracy |
|---|---|
| Goldani *et al.* [40] | 0.998 |
| Hakkak *et al.* [33] | 1.000 |
| Blackedge *et al.* [48] | 0.988 |
| Nasir *et al.* [42] | 0.990 |
| Proposed approach | **0.994** |

**TABLE II:** Comparison of test accuracy with previous works for dataset1

| Source | Method | Accuracy |
|---|---|---|
| [59] | LSTM | 0.82 |
| [60] | Bi-LSTM | 0.84 |
| [61] | Random Forest | 0.85 |
| [62] | Decision Trees | 0.92 |
| | Proposed approach | **0.97** |

**TABLE III:** Comparison of test accuracy with previous works for dataset2

| Authors | Accuracy |
|---|---|
| Ghanem *et al.* [37] | 0.48 |
| Singh *et al.* [32] | 0.87 |
| Ruchansky *et al.* [14] | 0.89 |
| Ahmed *et al.* [28] | 0.92 |
| Yang *et al.* [63] | 0.92 |
| O'Brien *et al.* [38] | 0.93 |
| Proposed approach | **0.99** |

**TABLE IV:** Comparison of test accuracy with previous works for dataset3

| Minibatch size | Accuracy | Precision | Recall | F1-score |
|---|---|---|---|---|
| 16 | 0.92 | 0.97 | 0.88 | 0.9226 |
| 32 | 0.95 | 0.95 | 0.94 | 0.9471 |
| 64 | 0.94 | 0.97 | 0.91 | 0.9420 |
| 128 | 0.93 | 0.89 | 0.98 | 0.9341 |
| 256 | 0.93 | 0.97 | 0.90 | 0.9309 |
| 512 | 0.94 | 0.96 | 0.93 | 0.9412 |
| 1024 | 0.93 | 0.97 | 0.90 | 0.9325 |

**TABLE V:** Accuracy, Precision, Recall, and F1-score for the unified dataset without preprocessing of input data

| Minibatch size | Accuracy | Precision | Recall | F1-score |
|---|---|---|---|---|
| 16 | 0.95 | 0.98 | 0.92 | 0.9511 |
| 32 | 0.96 | 0.98 | 0.94 | 0.9581 |
| 64 | 0.97 | 0.98 | 0.95 | 0.9666 |
| 128 | 0.96 | 0.97 | 0.96 | 0.9632 |
| 256 | 0.96 | 0.98 | 0.94 | 0.9606 |
| 512 | 0.95 | 0.97 | 0.94 | 0.9526 |
| 1024 | 0.95 | 0.95 | 0.94 | 0.9478 |

**TABLE VI:** Accuracy, Precision, Recall, and F1-score for the unified dataset with preprocessing of input data

The unified model from the three datasets is trained using the combined dataset and can be considered joint training. Results showed that the proposed training process performed well on the combined dataset than the other models, with 97% of accuracy. On the other hand, the Random Forest algorithm performed slightly worse than our proposed model with 91% of accuracy. Overall, the deep learning algorithms showed limited performance for all four datasets. However, deep learning models slightly improved the performance on average when trained on the combined dataset. Random Forest algorithm showed better performance on the three datasets, ex- cept it showed lower accuracy for dataset2. Also, the Logistic Regression model performed better on four datasets among the individual classifiers without showing less performance on any individual dataset. In the case of ensemble learners, Boosting Classifier performed very well on all four datasets with an average accuracy of 90%. Among the deep learning models, LSVM model [64] outperformed the CNN [35] and Bi-LSTM [35] models. Finally, the proposed approach showed better performance on over four datasets and improved 6% accuracy using the unified training strategy on the combined dataset than the Random Forest algorithm.

*G. Ablation study*

In this part, we will analyze the performance of the unified model by varying the number of encoder blocks. Table VIII shows the accuracy, precision, recall, and F1 scores by changing the number of encoder blocks from the selected pre-trained transformer model by varying the batch size. The alternate encoder blocks are removed at every step, and we recorded the performance results. We included the fifth encoder block in the model during the last phase and finetuned with similar conditions. Results indicate that the model with the best performance compared to other variations, with 1, 3, 5, 7, 9, and 11 layers with a minibatch size of 32. However, the model's performance significantly decreases when there is a reduction in the number of encoder blocks. However, using only the encoder block numbered 5 showed a lower accuracy of 83% for a batch size of 128. The obtained results can help the researchers design the model according to the hardware availability and user preferences. For example, if the hardware has the minimal capacity, the user can reduce the number of encoder blocks accordingly to run the model but incur a performance penalty.

| Model | Dataset1(Acc/Pre/Rec/F1) | Dataset2(Acc/Pre/Rec/F1) | Dataset3(Acc/Pre/Rec/F1) | Unified dataset (Acc/Pre/Rec/F1) |
|---|---|---|---|---|
| Machine Learning Models | | | | |
| Logistic Regression (LR) [21] | 0.97/0.98/0.98/0.9 | 0.91/0.92/0.90/0.91 | 0.91/0.93/0.92/0.92 | 0.87/0.88/0.86/0.87 |
| Linear SVM (LSVM) [21] | 0.98/0.98/0.98/0.98 | 0.37/0.31/0.32/0.32 | 0.53/0.54/1/0.7 | 0.86/0.88/0.86/0.87 |
| Multilayer Perceptron [21] | 0.98/0.97/1/0.98 | 0.35/0.32/0.36/0.34 | 0.94/0.93/0.96/0.95 | 0.90/0.92/0.88/0.90 |
| K-Nearest Neighbors (KNN) [21] | 0.99/0.91/0.87/0.99 | 0.28/0.22/0.24/0.23 | 0.82/0.85/0.81/0.83 | 0.77/0.80/0.74/0.77 |
| Random Forest (RF) [21] | 0.99/0.99/1/0.99 | 0.35/0.30/0.34/0.32 | 0.95/0.98/0.93/0.95 | 0.91/0.92/0.91/0.91 |
| Voting Classifier (RF, LR, KNN) [21] | 0.97/0.96/0.97/0.97 | 0.88/0.88/0.89/0.88 | 0.94/0.92/0.96/0.94 | 0.88/0.86/0.90/0.88 |
| Voting Classifier (LR,LSVM, CART) [21] | 0.96/0.94/0.97/0.96 | 0.86/0.86/0.87/0.86 | 0.92/0.88/0.96/0.92 | 0.85/0.83/0.89/0.86 |
| Bagging Classifier (decision trees) [21] | 0.988/0.98/0.98/0.9 | 0.92/0.92/0.93/0.92 | 0.92/0.92/0.92/0.92 | 0.86/0.86/0.86/0.86 |
| Boosting Classifier (AdaBoost) [21] | 0.98/0.99/0.99/0.99 | 0.94/0.94/0.94/0.94 | 0.94/0.96/0.94/0.95 | 0.89/0.92/0.89/0.90 |
| Boosting Classifier (XGBoost) [21] | 0.98/0.99/0.99/0.99 | 0.94/0.94/0.94/0.94 | 0.94/0.96/0.94/0.95 | 0.89/0.92/0.89/0.90 |
| Deep Learning Models | | | | |
| LSVM [21] [64] | 0.99/0.99/0.99/0.99 | 0.79/0.79/0.81/0.80 | 0.96/0.96/0.97/0.96 | 0.90/0.90/0.91/0.90 |
| CNN [21] [35] | 0.87/0.84/0.90/0.87 | 0.67/0.65/0.29/0.67 | 0.58/0.48/0.29/0.31 | 0.73/0.72/0.75/0.73 |
| Bi-LSTM [35] [21] | 0.86/0.92/0.78/0.84 | 0.52/0.43/0.59/0.44 | 0.57/0.50/0.35/0.35 | 0.62/0.65/0.61/0.57 |
| Our unified model (minibatch size) | **0.99/0.99/0.99/0.99** (32) | **0.97/0.96/0.99/0.97** (64) | **0.99/0.99/0.98/0.99** (128) | **0.97/0.98/0.95/0.97** (64) |

**TABLE VII:** Accuracy(Acc)/Precision(Pre)/Recall(Rec)/F1-score(F1) for the four datasets

| Encoder Blocks (mini batch size) | Accuracy | Precision | Recall | F1-score |
|---|---|---|---|---|
| 1,3,5,7,9,11 (16) | 0.94 | 0.97 | 0.92 | 0.9473 |
| 1,3,5,7,9,11 (32) | **0.96** | **0.94** | **0.96** | **0.9621** |
| 1,3,5,7,9,11 (64) | 0.95 | 0.97 | 0.93 | 0.9497 |
| 1,3,5,7,9,11 (128) | 0.95 | 0.97 | 0.92 | 0.9473 |
| 1,5,9 (16) | 0.94 | 0.98 | 0.91 | 0.9424 |
| 1,5,9 (32) | 0.95 | 0.99 | 0.91 | 0.9501 |
| 1,5,9 (64) | 0.96 | 0.98 | 0.93 | 0.9550 |
| 1,5,9 (128) | 0.94 | 0.99 | 0.89 | 0.9385 |
| 1,9 (16) | 0.94 | 0.96 | 0.92 | 0.9412 |
| 1,9 (32) | 0.95 | 0.97 | 0.93 | 0.9469 |
| 1,9 (64) | 0.94 | 0.96 | 0.92 | 0.9432 |
| 1,9 (128) | 0.94 | 0.94 | 0.93 | 0.9345 |
| 5 (16) | 0.87 | 0.83 | 0.95 | 0.8855 |
| 5 (32) | 0.90 | 0.96 | 0.85 | 0.8984 |
| 5 (64) | 0.90 | 0.88 | 0.93 | 0.9021 |
| 5 (128) | **0.83** | **0.88** | **0.78** | **0.8244** |

**TABLE VIII:** Accuracy, Precision, Recall, F1-Score for unified dataset by varying the number of encoder blocks with minibatch sizes

## V. FUTURE WORK

The classification tasks like fake review detection, sentiment analysis detection, mental health prediction using NLP, etc., with the combined multiple datasets, can use the unified training strategy [65, 66, 67]. Moreover, efficient hyperparameter tuning can be made for the unified model with a faster turnaround time with the help of data subset selection instead of using all the samples from individual datasets during the training process [68]. Furthermore, advanced hyperparameter tuning approaches can help to obtain the best performing unified models [69, 70]. A better-centralized model can be obtained using the unified training strategy with minor modifications in federated learning applications. The change includes giving equal weight values to accepted individual models [71]. In addition, the proposed strategy can be applied for healthcare applications by finetuning the differential private pretrained models can apply the unified training process on each client to obtain the best performing model [72, 73].

## VI. CONCLUSION

This manuscript proposed the unified training process for fine-tuning the pre-trained BERT transformer model using a combined dataset with the help of input preprocessing. The obtained model achieved 6% higher accuracy than the best-performing machine learning model like Random Forest Algorithm. Also, the input preprocessing technique helped reduce the training time up to 1.8× and improved performance by 2% on accuracy compared to the model trained without input preprocessing. To make the model compact, we further reduced the number of encoder blocks resulting in decreased performance. The proposed work can provide a future benchmark to train the models on individual datasets separately and can use the obtained information to formulate a unified model that can train on the combined dataset for the assigned classification task.

ACKNOWLEDGEMENT

We first thank the anonymous reviewers for their valuable comments that improved this paper. We then sincerely thank Dr. Liqun Shan for her valuable time, dedicated interest in giving suggestions, and efforts to restructure the paper. This work is supported in part by the US NSF under grants OIA-1946231 and CNS-2117785.